\documentclass[10pt,twocolumn,letterpaper]{article}

\usepackage{cvpr}              

\usepackage{graphicx}
\usepackage{amsmath}
\usepackage{amssymb}
\usepackage{booktabs}
\usepackage{subcaption} 
\usepackage{tikz}
\usepackage{footmisc}
\usepackage{amsfonts}
\usepackage{mathtools}
\usepackage{array}
\usepackage{tabularx}
\usepackage{textcomp}
\usepackage{pgfplots}
\usepackage{comment}

\newcolumntype{Y}{>{\centering\arraybackslash}X}

\definecolor{cvprblue}{rgb}{0.21,0.49,0.74}
\usepackage[pagebackref,breaklinks,colorlinks,allcolors=cvprblue]{hyperref}

\def\paperID{****} 
\def\confName{****}
\def\confYear{****}

\title{WiFlexFormer: Efficient WiFi-Based Person-Centric Sensing}

\author{Julian Strohmayer\qquad Matthias Wödlinger\qquad Martin Kampel\\
Computer Vision Lab, TU Wien\\
Favoritenstr. 9/193-1, 1040 Vienna, Austria\\
{\tt\small \{julian.strohmayer, matthias.woedlinger, martin.kampel\}@tuwien.ac.at}
}

\begin{document}
\maketitle
\begin{abstract}
We propose \textit{WiFlexFormer}, a highly efficient Transformer-based architecture designed for WiFi Channel State Information (CSI)-based person-centric sensing. We benchmark \textit{WiFlexFormer} against state-of-the-art vision and specialized architectures for processing radio frequency data and demonstrate that it achieves comparable Human Activity Recognition (HAR) performance while offering a significantly lower parameter count and faster inference times. With an inference time of just 10 ms on an Nvidia Jetson Orin Nano, \textit{WiFlexFormer} is optimized for real-time inference. Additionally, its low parameter count contributes to improved cross-domain generalization, where it often outperforms larger models. Our comprehensive evaluation shows that \textit{WiFlexFormer} is a potential solution for efficient, scalable WiFi-based sensing applications. The PyTorch implementation of \textit{WiFlexFormer} is publicly available at: \href{https://github.com/StrohmayerJ/WiFlexFormer}{https://github.com/StrohmayerJ/WiFlexFormer}.

\end{abstract}

\vspace{-3mm}
\section{Introduction}
\vspace{-2mm}
WiFi has emerged as a promising modality in person-centric sensing due to its advantages over optical approaches, including cost-effectiveness, unobtrusiveness, visual privacy protection, and the ability to perform long-range sensing through walls~\cite{YongsenWiFiSurvey2019, Fu234782379}. Together, these characteristics enable efficient, contactless monitoring of human activities in confined indoor environments without the need for per-room sensor deployment, representing a significant economic advantage ~\cite{StrohmayerICVS}. 

Channel State Information (CSI) serves as the foundation for modern WiFi-based person-centric sensing. CSI is a metric obtained in the Orthogonal Frequency-Division Multiplexing (OFDM) scheme, which subdivides a WiFi channel into multiple sub-channels with different carrier frequencies (subcarriers) \cite{HernandezWiFiOnTheEdgeSurvey}. This subdivision allows for fast, parallel transmission of data, while CSI provides detailed information about how each subcarrier is affected by the environment, enabling the correction of environment-induced noise at the receiver on a per-subcarrier basis, and through correlating the distinctive patterns of amplitude attenuation and phase shifts in CSI caused by specific human movements, applications such as Human Activity Recognition (HAR)~\cite{Liu45546465465464}.

While existing approaches to CSI-based HAR often rely on generic CNN-based vision architectures, they are not optimal due to their focus on local dependencies and shift-invariance~\cite{yang2023slnet}. To effectively leverage the unique properties of CSI, specialized architectures have been developed~\cite{ding2020rf,li2021two,yang2023slnet}. However, they tend to suffer from other problems such as overly complex designs and reliance on computationally expensive features, resulting in high inference times and limited practicality for real-time applications.

\textbf{Contributions.} To address these challenges, we make the following contributions: 
(I) We propose \textit{WiFlexFormer}, a Transformer-based architecture that achieves similar HAR performance with significantly lower parameter count and inference time, making it highly efficient and well-suited for WiFi-based real-time person-centric sensing applications. 
(II) We conduct comprehensive evaluations of \textit{WiFlexFormer} on publicly available WiFi datasets, assessing its HAR performance using amplitude and Doppler Frequency Shift (DFS) features and comparing it against existing state-of-the-art architectures. 
(III) We investigate the effectiveness of various subcarrier sub-sampling strategies to further optimize inference speed while maintaining model performance.

\vspace{-1mm}
\section{Related Work}
\vspace{-2mm}

Early works on learned methods for HAR using WiFi CSI data often relied on generic CNN-based vision architectures. Popular vision architectures such as \textit{ResNet}~\cite{he2016deep}, \textit{EfficientNet}~\cite{tan2021efficientnetv2}, and \textit{ShuffleNet}~\cite{ma2018shufflenet} have been widely adopted for WiFi sensing tasks due to their proven effectiveness in feature extraction and their adaptability to various input types. For instance, the SignFi system \cite{SignFiMa2018} employed a 9-layer CNN for sign language recognition, while Widar3.0~\cite{Zhang2022Widar} utilized a CNN-based architecture for cross-domain gesture recognition. Jiang et al.~\cite{jiang2018towards} proposed a CNN-based HAR framework that extracts features shared across different subjects and environments.

To better leverage the unique properties of CSI data, researchers have developed specialized architectures. \textit{SLNet} \cite{yang2023slnet} introduced a spectrogram learning neural network that uses complex-valued convolutional layers to extract features from DFS spectrograms. It relies on Principal Component Analysis (PCA) and Short-Time Fourier Transform (STFT) as preprocessing steps and introduces a submodule to remove spectral leakage due to STFT. While \textit{SLNet} achieves impressive performance across multiple WiFi sensing tasks, its main drawback is the computationally expensive preprocessing step, which may limit its applicability in real-time or resource-constrained scenarios.

Long Short-Term Memory (LSTM) networks have been utilized to capture temporal dependencies in CSI data \cite{shalaby2022utilizing}. Shi et al.~\cite{Shi2022} proposed an environment-robust device-free HAR system using CSI enhancement and one-shot learning to address the challenge of limited training data. Hybrid CNN-LSTM architectures have also been explored, such as in DeepSense~\cite{Zou2018DeepSenseDH}, which introduced an autoencoder long-term recurrent convolutional network for device-free HAR, combining the spatial feature extraction capabilities of CNNs with the temporal modeling of LSTMs.
Addressing the challenge of limited labeled data in new environments or for new users, \textit{RF-Net}~\cite{ding2020rf} introduced a meta-learning framework for one-shot human activity recognition using radio frequency (RF) signals such as WiFi. This approach aims to improve generalization across different domains and reduce the need for extensive data collection in each new setting.

More recently, Transformer-based architectures have been applied to WiFi sensing tasks, leveraging their ability to capture long-range dependencies and process sequential data efficiently. The Two-Stream Convolution Augmented Human Activity Transformer (\textit{THAT}) architecture \cite{li2021two} takes a different approach by utilizing a two-stream structure to capture both time-over-channel and channel-over-time features. Contrary to our method it utilizes two transformer encoders in parallel whose outputs are concatenated and processed for the prediction. Other works, such as MetaFi~\cite{yang2022metafi} for WiFi-based pose estimation and WiTransformer~\cite{yang2023witransformer} for gesture recognition, have further explored and adapted Transformer architectures for various WiFi sensing applications.
While these specialized architectures have shown promising results, they often suffer from high complexity and computational costs, limiting their practicality for real-time applications. Additionally, many existing approaches rely on computationally expensive feature extraction methods, further increasing inference times and energy consumption.

The proposed \textit{WiFlexFormer} addresses these limitations by using a lightweight Transformer-based architecture designed specifically for the efficient processing of WiFi CSI features. Unlike previous approaches that use generic vision architectures or overly complex specialized architectures, our method achieves comparable HAR performance with significantly lower parameter count and inference time. This makes it particularly well-suited for applications where low latency and energy efficiency are critical.

\begin{figure}[t!]
  \centering
   \includegraphics[width=0.725\linewidth]{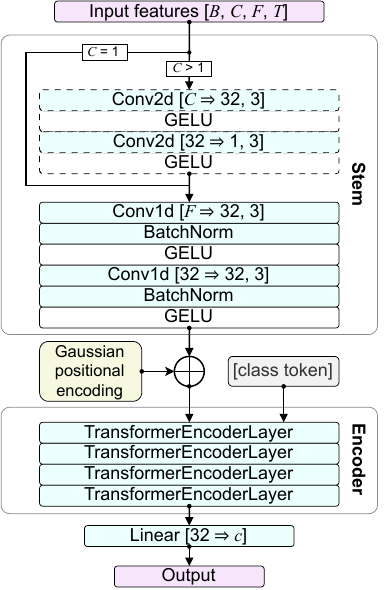}
   \vspace{-1mm}
   \caption{The proposed \textit{WiFlexFormer} architecture. Convolution parameters are denoted as: [input channels $\Rightarrow$ number of filters, kernel size]. The final linear layer has 32 input features and $c$ output features, the number of classes. Only the output at the position of the class token is used for the prediction, the remaining positions are discarded.}
   \label{figArchitecture}
   \vspace{-5mm}
\end{figure}

\vspace{-1mm}
\section{WiFlexFormer}
\vspace{-1mm}

The \textit{WiFlexFormer} architecture, illustrated in Figure ~\ref{figArchitecture}, comprises an initial stem module followed by a Transformer encoder. This design choice is motivated by the unique characteristics of WiFi CSI. The Transformer encoder has been shown to effectively capture long-range dependencies and global context \cite{VaswaniTransformer3854934580}, which is crucial for understanding the temporal and frequency patterns in CSI. The stem module provides initial feature extraction and dimensionality reduction, while the Transformer encoder enables the model to attend to relevant parts of the input sequence, regardless of their position.

\textbf{Input Features.} \textit{WiFlexFormer} expects generic real-valued input features in the shape [$B$, $C$, $F$, $T$], where $B$, $C$, $F$, and $T$ correspond to the batch, channel, frequency, and time dimensions, respectively. This allows for the processing of common WiFi features such as CSI, DFS, amplitude, phase, and their derivations. To handle complex-valued inputs like CSI, we follow the approach presented in ~\cite{yang2023slnet} by separating the real and imaginary parts and storing them in two real-valued channels. This results in an input tensor shape of [$B$, 2, $F$, $T$]. For unprocessed inputs such as CSI or amplitude, the dimensionality of $F$ corresponds to the number of subcarriers, while for features resulting from STFT, such as DFS, $F$ corresponds to the number of frequency bins.

\textbf{Stem.} The stem of \textit{WiFlexFormer} is designed to handle various input features and perform initial feature extraction and dimensionality reduction. It consists of two main components: a 2D stem for multi-channel inputs and a 1D stem for further processing. For inputs with multiple channels ($C > 1$), such as DFS features where each subcarrier generates a 2D spectrogram, we first apply a 2D stem comprising two convolutional layers with kernel size $(1, 3)$, the first maintaining the number of input channels and the second reducing it to 1, both using GELU activation functions.

Following the 2D stem (or directly for single-channel inputs like amplitude features), we apply a 1D stem consisting of two blocks, each containing a 1D convolutional layer (kernel size 3, 32 filters), batch normalization, GELU activation, and dropout (rate 0.1). This 1D stem reduces the frequency dimension from its input size to a fixed dimension of 32. 

This flexible architecture allows \textit{WiFlexFormer} to handle various input types: amplitude features $[B, 1, F, T]$ ($F$ is the number of subcarriers) and DFS features $[B, C, F, T]$ ($C$ is the number of subcarriers, $F$ is the number of frequency bins). The stem's design replaces heuristic preprocessing steps, providing an end-to-end learnable approach for feature extraction and noise reduction. The temporal receptive field of 5 in the 1D stem helps accumulate information from adjacent positions, enhancing the model's ability to handle noisy inputs.

\textbf{Positional Encoding and Class Token.} To encode the temporal dimension, we apply Gaussian positional encoding to the stem's output, following the method described by Li et al.~\cite{li2021two}. The resulting encoded features are combined with a class token before being input into a four-layer Transformer encoder. This token serves as an aggregator of global information, and its output will be used for the final classification. The use of a class token rather than direct feature aggregation, as in the work from Li et al.~\cite{li2021two}, prevents blurring temporal relationships.

\textbf{Encoder.} Each layer of the encoder contains 16 attention heads and a feedforward dimension of 64. The final prediction is generated by processing the class token output through a linear classification head. The \textit{WiFlexFormer} can be trained end-to-end and remains relatively lightweight, containing only $\approx$ 50k parameters (depending on the input shape). The proposed architecture is designed to work directly with CSI amplitude features and, therefore, does not rely on complicated or slow feature extraction methods.

\textbf{Cross-Domain Generalization.} \textit{WiFlexFormer} incorporates several features that contribute to its potential for improved cross-domain generalization compared to other architectures. Its flexible input handling allows the processing of various features without extensive preprocessing, adapting to different data representations across domains. The combination of convolutional layers in the stem and the subsequent Transformer encoder allows for multi-scale feature learning, capturing generalizable patterns at various levels of abstraction. With only about 50k parameters, \textit{WiFlexFormer}'s lightweight design acts as a form of regularization, potentially preventing overfitting to domain-specific nuances.

\begin{figure*}[ht!]
     \centering
     \begin{subfigure}[b]{0.35\textwidth}
         \centering
         \includegraphics[width=\textwidth]{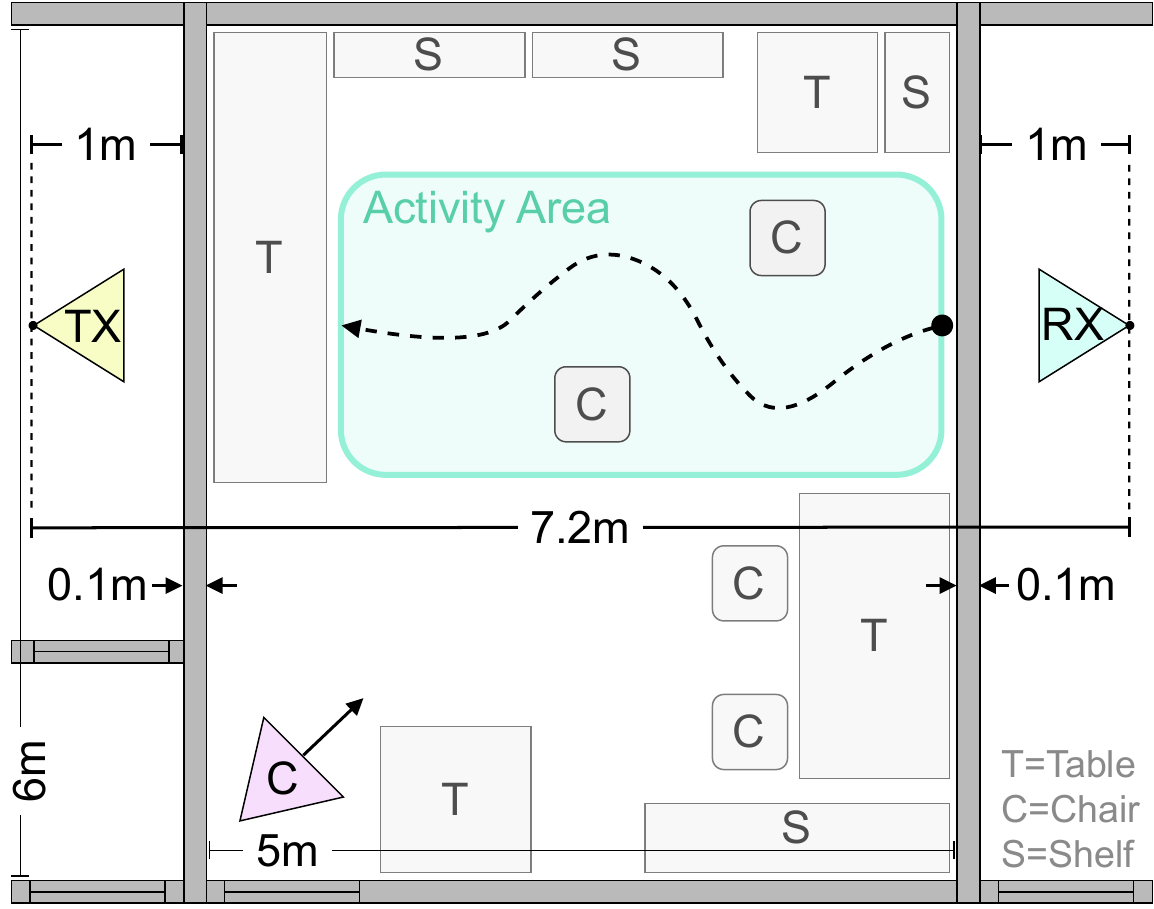}
         \caption{3DO setup on days 1 and 2.}
          \label{fig3DOday12Setup}
     \end{subfigure}
     \hfill
     \begin{subfigure}[b]{0.35\textwidth}
         \centering
         \includegraphics[width=\textwidth]{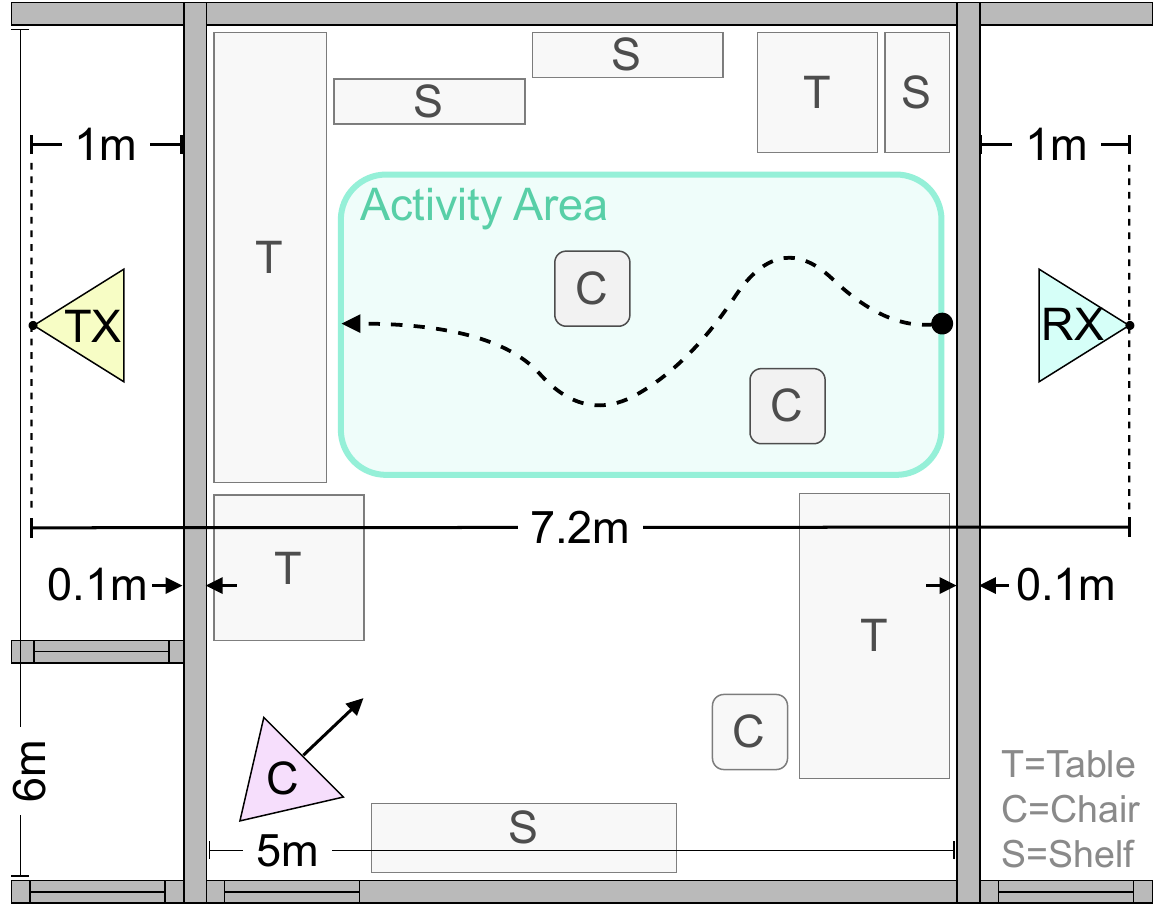}
         \caption{3DO setup on day 3.}
          \label{fig3DOday3Setup}
     \end{subfigure}
     \hfill
     \begin{subfigure}[b]{0.275\textwidth}
         \centering
         \includegraphics[width=\textwidth]{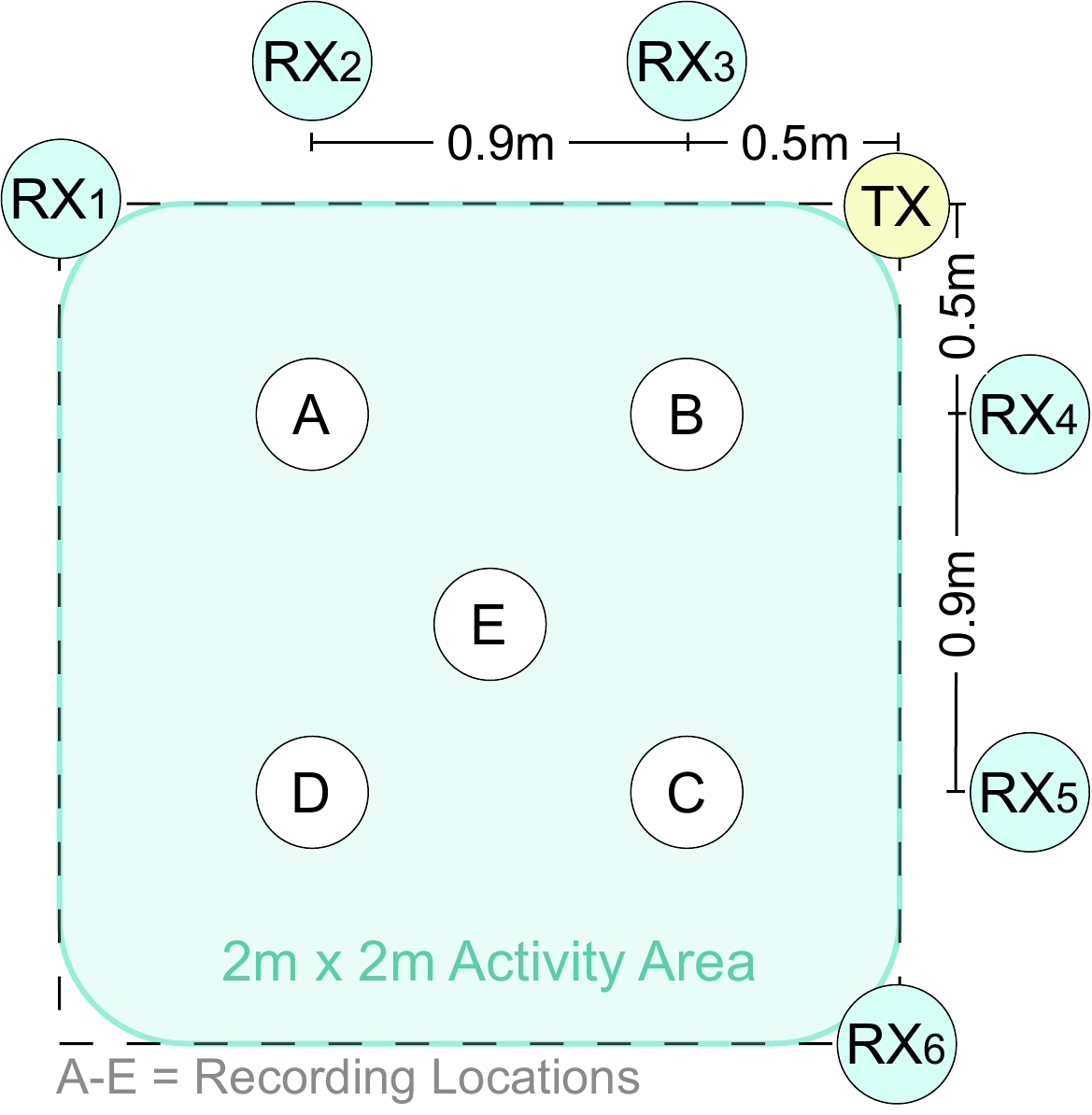}
         \caption{Widar3.0 setup.}
          \label{figWidar3Setup}
     \end{subfigure}
        \vspace{-1mm}
       \caption{(a-b) 3DO Recording setup over three consecutive days: (a) setup on days 1 and 2, and (b) setup on day 3, featuring static environmental variations due to furniture rearrangement. The transmitter-receiver arrangement and the designated activity area remain fixed throughout the experiment. (c) Widar3.0 recording setup featuring a single transmitter and six receivers.}
       \label{figureEnvironment}
       \vspace{-3mm}
\end{figure*}

\vspace{-1mm}
\section{Evaluation}
\vspace{-1mm}
The WiFi-based person-centric sensing capabilities of \textit{WiFlexFormer} are evaluated on two publicly available datasets. These datasets encompass a variety of systems, transmitter-receiver configurations, and recording environments and include both micro- and macroscopic human activities. Additionally, our evaluation covers diverse scenarios, such as Line-of-Sight (LoS) and through-wall, i.e., Non-Line-of-Sight (NLoS), sensing. Given that cross-domain generalization remains a significant challenge in WiFi-based sensing~\cite{chen2023cross}, our evaluation is structured to measure performance in both in-domain and cross-domain contexts. To provide a comprehensive assessment, \textit{WiFlexFormer} is compared against a range of state-of-the-art vision architectures, as well as architectures specifically designed for processing RF signals, such as WiFi. The evaluation metrics include both HAR performance and inference speed. Finally, to optimize inference speed, we explore various subcarrier sub-sampling strategies using amplitude and DFS features. DFS features are investigated alongside amplitude features since they are a popular choice among existing methods~\cite{yang2023slnet, ding2020rf} and due to their potential robustness to environmental changes. However, DFS features require more computational resources for extraction.

\subsection{Data}
\paragraph{3DO Dataset.}
The 3DO dataset is a WiFi through-wall HAR dataset specifically designed to evaluate model cross-domain generalization under varying static, dynamic, and temporal environmental conditions, making it distinct from existing datasets, which generally focus on multi-subject scenarios or same-room data collection. The dataset introduces a clear separation between environmental variations and participant-specific variations, focusing solely on environmental factors for isolated generalization assessment in a through-wall scenario. Unlike many other datasets that include multiple participants, 3DO uses a single participant to eliminate physiological variation, a well-known domain variation already covered by existing datasets. This approach allows for a focused evaluation of static and dynamic environmental changes in a through-wall scenario.

The dataset consists of recordings of three macroscopic activities: \textit{walking}, \textit{sitting}, and \textit{lying}, performed in an office environment over three consecutive days. Figure \ref{fig3DOday12Setup} shows the recording setup, where activities occur in a central room (6 m $\times$ 5 m) furnished with standard office items. Adjacent rooms house the WiFi transmitter and receiver, spaced 7.2 m apart and separated by 0.1 m thick plasterboard walls. The hardware setup remains fixed across all days. The dataset introduces three types of variation across days. On day 1, the participant performs the activities in a static environment, establishing a baseline for in-domain performance. Day 2 introduces dynamic variations as the participant adjusts movement patterns, and temporal variations arise from WiFi hardware~\cite{Lee10185958}, representing a mild cross-domain challenge. Day 3 adds static environmental variation by rearranging furniture, along with continued dynamic and temporal variations, creating a more complex cross-domain scenario for generalization (Figure \ref{fig3DOday3Setup}).

Data collection uses the 2.4 GHz WiFi system by Strohmayer et al.~\cite{StrohmayerICLR}, with an \textit{ESP32-S3-DevKitC-1U}, \textit{ESP32-S3-WROOM-1U} microcontroller, and \textit{ALFA Network APA-M25} antenna. WiFi packets are transmitted at 100 Hz and captured with synchronized video for activity labeling. Each day, five 5-minute sequences are recorded per activity. In \textit{walking}, the participant moves freely within the area; in \textit{sitting}, they alternate between two chairs with random head, arm, and leg movements; in \textit{lying}, a fall detection scenario is simulated. In total, the dataset contains over 1.2 million labeled WiFi packets and is publicly available\footnote{3DO Dataset, \href{https://zenodo.org/records/10925351}{https://zenodo.org/records/10925351}\label{footnoteDataset}}. The distribution of samples across days and activity classes is shown in Table \ref{tableDataset}.

\begin{table}[t!]
\centering
\begin{tabularx}{0.475\textwidth}{>{\arraybackslash}p{12mm}>{\raggedleft\arraybackslash}p{12mm}>{\raggedleft\arraybackslash}p{12mm}>{\raggedleft\arraybackslash}p{12mm}>{\raggedleft\arraybackslash}p{15mm}}
\toprule
\small{Day} & \small{\textit{walking}} & \small{\textit{sitting}} & \small{\textit{lying}} & \small{all classes}\\
\midrule
\small{1} & \small{105,288} & \small{142,242} & \small{149,803} & \small{397,333}\\
\small{2} & \small{110,146} & \small{181,991} & \small{181,332} & \small{473,469}\\
\small{3} & \small{152,504} & \small{150,338} & \small{87,935} & \small{390,777}\\
\midrule
 \small{Total} & \small{367,938} & \small{474,571} & \small{419,070} & \small{1,261,579}\\
\bottomrule
\end{tabularx}
\vspace{-1mm}
\caption{Distribution of data samples (WiFi packets) across days and activity classes: \textit{walking}, \textit{sitting}, and \textit{lying} in the 3DO dataset.}
\vspace{-5mm}
\label{tableDataset}
\end{table}

\vspace{-3mm}
\paragraph{Widar3.0-G6 Dataset.} The Widar3.0 dataset \cite{Zhang2022Widar} is one of the most widely used person-centric WiFi datasets, featuring CSI recordings of 22 human hand gestures performed by 16 participants in three different indoor environments. Since not all 22 gestures are consistently performed in all three environments, a subset of Widar3.0, referred to as Widar3.0-G6~\cite{Hou2024RFBoostUA}, is often utilized instead of the full dataset. This subset includes 6 gestures that are performed across all three environments by 15 users, resulting in a total of 11,250 hand gesture samples. The recording setup, shown in Figure \ref{figWidar3Setup}, consists of one 5.825 GHz WiFi transmitter (TX) and six receivers (RX$_{n}$), each equipped with an \textit{Intel WiFi Link 5300} wireless NIC that has three antennas. The CSI of 90 subcarriers (3 antennas $\times$ 30 subcarriers) is collected at each receiver using the Linux CSI Tool~\cite{Halperin43578934578}, utilizing a packet sending rate of 1,000 Hz.

\vspace{-2mm}
\subsection{Model Training}
\vspace{-1mm}
\paragraph{Architectures.} We compare the performance of \textit{WiFlexFormer} against a range of CNN-based vision architectures and specialized architectures for the processing of RF signals such as WiFi. The vision architectures include \textit{EfficientNetV2s}~\cite{tan2021efficientnetv2}, \textit{ResNet18}~\cite{he2016deep}, and \textit{ShuffleNetV2x0.5}~\cite{ma2018shufflenet}, while the RF signal processing architectures include \textit{RF-Net}~\cite{ding2020rf}, \textit{SLNet}~\cite{yang2023slnet}, both of which rely on DFS features and \textit{THAT}\cite{li2021two}, which expects amplitude features as input.

\vspace{-1mm}
\paragraph{Training Data.} For model training, we utilize the 3DO dataset and the Widar3.0-G6 dataset, employing both CSI amplitude and DFS features. For the 3DO dataset, we perform a 3:1:1 split on day 1 data for training, validation, and testing to evaluate in-domain performance. Data from days 2 and 3 are reserved for testing cross-domain generalization under dynamic and static environmental variations, respectively. We use the CSI of the 52 Legacy Long Training Field (L-LTF) subcarriers, with samples extracted over a window of 351 WiFi packets, equivalent to 3.51 seconds at a 100 Hz packet sending rate, a duration determined empirically.

The Widar3.0-G6 dataset is employed to assess cross-receiver generalization. It is split into two subsets: one containing data from receivers RX$_{1-3}$, used for training with an 8:2 training-validation split, and the other from receivers RX$_{4-6}$, reserved for testing. To align with the single-link nature of the 3DO dataset, only the CSI from antenna 1 at each receiver is used, resulting in a selection of 30 subcarriers. Temporal sub-sampling is performed at 100 Hz, with the sampling window length set to 369 packets based on the longest sample length post-sub-sampling, while shorter samples are zero-padded to this length.

Both amplitude and DFS features are extracted from CSI data. DFS features are computed on a per-subcarrier basis using STFT with a Gaussian window and a segment and FFT length of 125 WiFi packets. A frequency band-pass filter from -60 Hz to 60 Hz, as proposed in ~\cite{yang2023slnet}, is applied, resulting in 121 frequency bins. The input shapes for amplitude features are [$B$, 1, 52, 351] for the 3DO dataset and [$B$, 1, 30, 369] for the Widar3.0-G6 dataset. For DFS features, the input shapes are [$B$, 52, 121, 351] and [$B$, 30, 121, 369] for the 3DO and Widar3.0-G6 datasets, respectively. Furthermore, for \textit{SLNet}, real and imaginary parts of the complex-valued DFS features are stored separately in an additional dimension. The remaining models are fed with the absolute value of the computed DFS features. 

\vspace{-3mm}
\paragraph{Training Hyperparameters.}
For the activity recognition task, each architecture and feature configuration is trained from scratch in three independent runs with different random seeds, over 10 epochs. We use the AdamW optimizer with a learning rate of 1e-3 and a weight decay of 1e-3, optimizing for cross-entropy loss. To address class imbalances in the datasets, a balanced random sampler is employed. Training is conducted with a batch size of 32, and no data augmentation is applied, allowing us to evaluate the stand-alone generalization performance of each architecture. For each run, the best model, with respect to validation loss, is selected for evaluation on the test sets.

\vspace{-1mm}
\subsection{Results}
We evaluate model performance using standard metrics such as recall, precision, F1-score, and accuracy, computed on the test datasets for each model. To account for variability between runs, we report the mean and standard deviation of these metrics across three independent training runs, providing a more robust performance measure. Due to spatial constraints, only accuracy measurements are presented in the following; a detailed summary of all metrics is provided in the supplementary material. 

\vspace{-3mm}
\paragraph{Inference Time.}
To provide context for the HAR performance results, we first evaluate the inference times of all models, which reflect their parameter count and computational efficiency. We measure the inference time using amplitude and DFS features on an Nvidia Jetson Orin Nano single-board computer with 8 GB of VRAM. A batch size of 1 is used, resulting in input shapes of [1, 1, 52, 351] and [1, 52, 121, 351] for amplitude and DFS features, respectively. For each configuration, we perform 100 warm-up iterations preceding the measurement of the mean inference time over 1,000 iterations. The results, presented in Table \ref{tableInferenceTime}, show that \textit{WiFlexFormer} achieves the lowest inference times for both feature types, with a mean inference time of 9.26 ms for amplitude features and 11.06 ms for DFS features. The second-fastest model, \textit{ResNet18}, achieves mean inference times of 9.67 ms for amplitude features and 12.01 ms for DFS features. In comparison, specialized models such as \textit{RF-Net} (427.20 ms for DFS), \textit{SLNet} (322.31 ms for DFS), and \textit{THAT} (37.87 ms for amplitude) exhibit significantly higher inference times due to their larger parameter counts. These findings demonstrate that \textit{WiFlexFormer} provides more efficient inference across both feature types, making it particularly suitable for real-time edge applications where low latency is essential.

\begin{table}[t!]
\centering
\begin{tabularx}{0.475\textwidth}{>{\arraybackslash}p{30mm}>{\raggedleft\arraybackslash}Y>{\raggedleft\arraybackslash}Y}
\toprule
& \multicolumn{2}{c}{Inference Time [ms]} \\ \cmidrule(lr){2-3}
Model & Amplitude & DFS \phantom{xx.}\\
\midrule
RF-Net~\cite{ding2020rf} & - & 427.20 \scriptsize{$\pm$ 13.6}\\ 
SLNet~\cite{yang2023slnet} & - & 322.31 \scriptsize{$\pm$ 7.25} \\
EfficientNetV2s~\cite{tan2021efficientnetv2} & 67.72 \scriptsize{$\pm$ 0.64} & 68.66 \scriptsize{$\pm$ 0.78} \\
THAT~\cite{li2021two} & 37.87 \scriptsize{$\pm$ 0.49} & - \\
ShuffleNetV2x0.5~\cite{ma2018shufflenet} & 22.16 \scriptsize{$\pm$ 0.55} & 22.59 \scriptsize{$\pm$ 0.50} \\
ResNet18~\cite{he2016deep} & 9.67 \scriptsize{$\pm$ 0.19} & 12.01 \scriptsize{$\pm$ 0.42} \\
WiFlexFormer [ours] & \underline{9.26} \scriptsize{$\pm$ 0.39} & \underline{11.06} \scriptsize{$\pm$ 0.63} \\
\bottomrule
\end{tabularx}
\vspace{-2mm}
\caption{Inference time comparison between models for amplitude and DFS features. Inference time is reported as the mean inference time over 1,000 iterations (excluding 100 warm-up iterations) on an \textbf{Nvidia Jetson Orin Nano} using a batch size of 1.}
\label{tableInferenceTime}
\vspace{-6mm}
\end{table}

\vspace{-3mm}
\paragraph{HAR Performance on 3DO.} Table \ref{tableResults3DOAmplitude} presents the in- and cross-domain activity recognition performance for all models on the 3DO dataset using amplitude features. Day 1 represents in-domain performance, while days 2 and 3 reflect cross-domain performance under dynamic and static environmental variations, respectively.

For in-domain performance (day 1), all models perform similarly, with vision-based models such as \textit{ResNet18} and \textit{ShuffleNetV2x0.5} slightly outperforming specialized models. However, \textit{WiFlexFormer} achieves a competitive accuracy of 98.41\%, outperforming the specialized model \textit{THAT}, while using only a fraction of the parameters (0.05 M vs. 7.96 M).

In the cross-domain evaluation on day 2, which introduces dynamic variations, \textit{WiFlexFormer} demonstrates strong generalization capabilities, achieving 85.26\% accuracy, second only to \textit{THAT} at 88.03\%. In contrast, vision-based models show a noticeable drop in accuracy, with \textit{ResNet18} reaching only 83.07\%.

On day 3, which adds the challenge of static environmental variation, \textit{WiFlexFormer} outperforms all other models with an accuracy of 86.98\%, highlighting its robustness in challenging cross-domain scenarios. Notably, \textit{THAT} experiences a significant drop in accuracy to 75.61\%, while vision-based models like \textit{ResNet18} struggle to maintain performance, achieving only 78.70\%.

Overall, considering both in-domain and cross-domain performance, \textit{WiFlexFormer}, using amplitude features, emerges as the best-performing model, offering superior generalization at a dramatically lower parameter count (0.05 M) compared to models like \textit{EfficientNetV2s} (20.18 M) and \textit{ResNet18} (11.17 M), making it highly efficient for real-world WiFi-based HAR applications.

\begin{table}[t!]
\centering
\begin{tabularx}{0.475\textwidth}{>{\arraybackslash}p{32mm}>{\raggedleft\arraybackslash}p{15mm}>{\centering\arraybackslash}p{3mm}Y}
\toprule
Model & Parameters & D & Accuracy $\uparrow$ \\
\midrule
EfficientNetV2s~\cite{tan2021efficientnetv2} & 20.18 M & 1 & 99.11 \scriptsize{$\pm$ 0.06} \\
ResNet18~\cite{he2016deep} & 11.17 M & 1 & \underline{99.38} \scriptsize{$\pm$ 0.10} \\
THAT~\cite{li2021two} & 7.96 M & 1 & 98.01 \scriptsize{$\pm$ 0.59} \\
ShuffleNetV2x0.5~\cite{ma2018shufflenet} & 0.34 M & 1 & 99.36 \scriptsize{$\pm$ 0.08} \\
WiFlexFormer [ours] & 0.05 M & 1 & 98.41 \scriptsize{$\pm$ 0.66} \\
\midrule
EfficientNetV2s~\cite{tan2021efficientnetv2} & 20.18 M & 2 & 80.01 \scriptsize{$\pm$ 2.18} \\
ResNet18~\cite{he2016deep} & 11.17 M & 2 & 83.07 \scriptsize{$\pm$ 0.26} \\
THAT~\cite{li2021two} & 7.96 M & 2 & \underline{88.03} \scriptsize{$\pm$ 2.77} \\
ShuffleNetV2x0.5~\cite{ma2018shufflenet} & 0.34 M & 2 & 80.25 \scriptsize{$\pm$ 2.52} \\
WiFlexFormer [ours] & 0.05 M & 2 & 85.26 \scriptsize{$\pm$ 3.56} \\
\midrule
EfficientNetV2s~\cite{tan2021efficientnetv2} & 20.18 M & 3 & 78.44 \scriptsize{$\pm$ 2.32} \\
ResNet18~\cite{he2016deep} & 11.17 M & 3 & 78.70 \scriptsize{$\pm$ 1.05} \\
THAT~\cite{li2021two} & 7.96 M & 3 & 75.61 \scriptsize{$\pm$ 0.32} \\
ShuffleNetV2x0.5~\cite{ma2018shufflenet} & 0.34 M & 3 & 70.49 \scriptsize{$\pm$ 5.87} \\
WiFlexFormer [ours] & 0.05 M & 3 & \underline{86.98} \scriptsize{$\pm$ 2.94} \\
\bottomrule
\end{tabularx}
\vspace{-2mm}
\caption{In- and cross-domain activity recognition performance on the \textbf{3DO dataset} using \textbf{amplitude features}. D indicates the day of data collection (see Table \ref{tableDataset}). All models are trained on day 1 with a 3:1:1 training-validation-test split. Amplitude features from all 52 subcarriers are used as input. Results are presented as mean and standard deviation across three independent runs with random initialization.}
\label{tableResults3DOAmplitude}
\vspace{-5mm}
\end{table}

\begin{table}[t!]
\centering
\begin{tabularx}{0.475\textwidth}{>{\arraybackslash}p{32mm}>{\raggedleft\arraybackslash}p{14mm}>{\centering\arraybackslash}p{3mm}Y}
\toprule
Model & Parameters & D & Accuracy $\uparrow$ \\
\midrule
RF-Net~\cite{ding2020rf} & 349.57 M & 1 & 88.19 \scriptsize{$\pm$ 1.00} \\
SLNet~\cite{yang2023slnet} & 146.27 M & 1 & 87.50 \scriptsize{$\pm$ 4.29} \\
EfficientNetV2s~\cite{tan2021efficientnetv2} & 20.19 M & 1 & \underline{97.67} \scriptsize{$\pm$ 0.03} \\
ResNet18~\cite{he2016deep} & 11.33 M & 1 & 94.92 \scriptsize{$\pm$ 1.00} \\
ShuffleNetV2x0.5~\cite{ma2018shufflenet} & 0.36 M & 1 & 97.48 \scriptsize{$\pm$ 0.40} \\
WiFlexFormer [ours] & 0.06 M & 1 & 92.83 \scriptsize{$\pm$ 0.37} \\
\midrule
RF-Net~\cite{ding2020rf} & 349.57 M & 2 & 62.89 \scriptsize{$\pm$ 2.07} \\
SLNet~\cite{yang2023slnet} & 146.27 M & 2 & 63.87 \scriptsize{$\pm$ 12.2} \\
EfficientNetV2s~\cite{tan2021efficientnetv2} & 20.19 M & 2 & \underline{86.58} \scriptsize{$\pm$ 3.34} \\
ResNet18~\cite{he2016deep} & 11.33 M & 2 & 75.16 \scriptsize{$\pm$ 0.08} \\
ShuffleNetV2x0.5~\cite{ma2018shufflenet} & 0.36 M & 2 & 71.21 \scriptsize{$\pm$ 1.19} \\
WiFlexFormer [ours] & 0.06 M & 2 & 79.91 \scriptsize{$\pm$ 1.42} \\
\midrule
RF-Net~\cite{ding2020rf} & 349.57 M & 3 & 59.58 \scriptsize{$\pm$ 1.63} \\
SLNet~\cite{yang2023slnet} & 146.27 M & 3 & \underline{76.91} \scriptsize{$\pm$ 7.80} \\
EfficientNetV2s~\cite{tan2021efficientnetv2} & 20.19 M & 3 & 73.62 \scriptsize{$\pm$ 4.52} \\
ResNet18~\cite{he2016deep} & 11.33 M & 3 & 69.30 \scriptsize{$\pm$ 7.05} \\
ShuffleNetV2x0.5~\cite{ma2018shufflenet} & 0.36 M & 3 & 71.47 \scriptsize{$\pm$ 3.54} \\
WiFlexFormer [ours] & 0.06 M & 3 & 74.18 \scriptsize{$\pm$ 3.15} \\
\bottomrule
\end{tabularx}
\vspace{-2mm}
\caption{In- and cross-domain activity recognition performance on the \textbf{3DO dataset} using \textbf{DFS features}. D indicates the day of data collection (see Table \ref{tableDataset}). All models are trained on day 1 with a 3:1:1 training-validation-test split. Amplitude features from all 52 subcarriers are used as input. Results are presented as mean and standard deviation across three independent runs with random initialization.}
\label{tableResults3DODFS}
\vspace{-5mm}
\end{table}

Table \ref{tableResults3DODFS} shows the in- and cross-domain HAR performance for all models on the 3DO dataset using DFS features. While \textit{WiFlexFormer} is outperformed by larger vision models such as \textit{EfficientNetV2s} and \textit{ShuffleNetV2x0.5}, this is expected due to their high parameter count, which allows them to make better use of the dense DFS features across all subcarriers. In contrast, \textit{WiFlexFormer} prioritizes strong feature compression in its design to remain computationally efficient, which limits its ability to leverage the full richness of DFS data. Despite this, \textit{WiFlexFormer} delivers a competitive in-domain accuracy of 92.83\%.

In cross-domain evaluations, especially on day 2, \textit{WiFlexFormer} demonstrates reasonable generalization, achieving 79.91\% accuracy, outperforming larger models such as \textit{ResNet18} and \textit{ShuffleNetV2x0.5}, and coming close to \textit{EfficientNetV2s}. On day 3, which introduces static environmental variations, \textit{WiFlexFormer} continues to show stability, with an accuracy of 74.18\%, higher than \textit{ResNet18} and comparable to other models, while \textit{SLNet} and \textit{EfficientNetV2s} exhibit a larger drop in performance.

One notable observation is the high run-to-run variance exhibited by the larger models, such as \textit{SLNet}, across all days, indicating instability during training, especially when faced with out-of-distribution samples. In contrast, \textit{WiFlexFormer} consistently shows lower variance, suggesting that its low parameter count may provide a natural regularization effect, leading to more stable training and performance across different domains.

Interestingly, none of the models, including the high-parameter models, were able to fully leverage DFS features for HAR, as the overall performance with DFS is notably lower than with amplitude features. This suggests that DFS features may not be optimal for this through-wall sensing scenario. We hypothesize that the noise induced by complex signal scattering and the highly noisy phase information involved in DFS computation likely contribute to the lower performance.

From an efficiency perspective, this outcome is promising: amplitude features, which have a lower dimensionality and do not require computationally expensive preprocessing, outperform DFS features requiring computationally expensive preprocessing, such as STFT, across all models. For \textit{WiFlexFormer}, this is particularly advantageous, as it achieves better HAR performance with simpler, faster-to-process amplitude features, making it an ideal solution for WiFi-based real-time sensing applications.

\begin{table}[t!]
\centering
\begin{tabularx}{0.475\textwidth}{>{\arraybackslash}p{32mm}>{\raggedleft\arraybackslash}p{18mm}Y}
\toprule
Model & Parameters & Accuracy $\uparrow$ \\
\midrule
EfficientNetV2s~\cite{tan2021efficientnetv2} & 20.18 M & \underline{51.98} \scriptsize{$\pm$ 0.45} \\
ResNet18~\cite{he2016deep} & 11.17 M & 51.38 \scriptsize{$\pm$ 1.84} \\
THAT~\cite{li2021two} & 8.48 M & 49.84 \scriptsize{$\pm$ 0.62} \\
ShuffleNetV2x0.5~\cite{ma2018shufflenet} & 0.35 M & 51.74 \scriptsize{$\pm$ 0.84} \\
WiFlexFormer [ours] & 0.04 M & 49.38 \scriptsize{$\pm$ 0.40} \\
\bottomrule
\end{tabularx}
\vspace{-2mm}
\caption{Cross-receiver gesture recognition performance on the \textbf{Widar3.0-G6 dataset} using \textbf{amplitude features}. All models are trained on data from receivers 1-3 with an 8:2 training-validation split and tested on receivers 4-6. Amplitude features from all 30 subcarriers are used as input. Results are presented as mean and standard deviation across three independent runs with random initialization.}
\label{tableWIDAR3G6Amplitude}
\vspace{-5mm}
\end{table}

\vspace{-3mm}
\paragraph{HAR Performance on Widar3.0-G6.}  
Table \ref{tableWIDAR3G6Amplitude} presents the cross-receiver gesture recognition performance using amplitude features on the Widar3.0-G6 dataset. Overall, all models perform similarly, with \textit{EfficientNetV2s} achieving the highest accuracy of 51.98\%. Vision-based models generally outperform specialized models such as \textit{THAT} and \textit{WiFlexFormer}, but the performance differences remain small. For instance, \textit{WiFlexFormer} trails \textit{EfficientNetV2s} by only 2.6\% accuracy, despite the latter having a 500x larger parameter count, demonstrating that \textit{WiFlexFormer} offers competitive accuracy with a much smaller model.

The results using DFS features, as shown in Table \ref{tableWIDAR3G6DFS}, tell a similar story. \textit{EfficientNetV2s} again achieves the highest accuracy with 51.34\%, outperforming specialized models like \textit{RF-Net} (48.11\%) and \textit{SLNet} (50.63\%). \textit{WiFlexFormer} delivers a competitive accuracy of 49.72\%, trailing \textit{EfficientNetV2s} by only 1.62\%. Notably, \textit{WiFlexFormer} outperforms \textit{RF-Net} by 1.61\% and comes close to \textit{SLNet}, despite their significantly larger parameter counts.

Overall, the performance across models on the Widar3.0-G6 dataset is quite close, and we observe no significant advantage in using DFS features over amplitude features in terms of cross-receiver generalization. Given that amplitude features require no computationally expensive preprocessing or parameter tuning, they remain a more efficient option for cross-domain generalization tasks.

\vspace{-2mm}
\paragraph{Subcarrier Selection.} As an additional experiment, we evaluate the effectiveness of different subcarrier sub-sampling strategies using the 3DO dataset. While a straightforward way to process CSI-based features without information loss would be to pass features from all subcarriers to a model, this can result in high computational complexity and slow inference. This is particularly true for DFS features, which require STFT preprocessing on a per-subcarrier basis, incurring high computational costs. As demonstrated in~\cite{Hou2024RFBoostUA}, it is neither efficient nor necessary to consider all subcarriers, as inter-subcarrier information, while complementary, is highly correlated. Utilizing a smart subcarrier sub-sampling strategy that eliminates redundant subcarriers can reduce computational costs, shorten inference times, and result in similar model accuracy compared to an approach using all subcarriers. To this end, we evaluate the effectiveness of different subcarrier sub-sampling strategies, including random sampling from all subcarriers, band-restricted random sampling, uniform sampling, and projection-based sampling utilizing PCA for dimensionality reduction. The results of our evaluation for amplitude and DFS features are presented in Figures \ref{figSSAmplitude} and \ref{figSSDFS}, respectively. The presented accuracy measurements represent the mean and standard deviation across days 1–3, thus capturing both in-domain and cross-domain HAR performance.

\begin{table}[t!]
\centering
\begin{tabularx}{0.475\textwidth}{>{\arraybackslash}p{32mm}>{\raggedleft\arraybackslash}p{18mm}Y}
\toprule
Model & Parameters & Accuracy $\uparrow$ \\
\midrule
RF-Net~\cite{ding2020rf} & 120.58 M & 48.11 \scriptsize{$\pm$ 0.37} \\
SLNet~\cite{yang2023slnet} & 88.88 M & 50.63 \scriptsize{$\pm$ 0.02} \\
EfficientNetV2s~\cite{tan2021efficientnetv2} & 20.19 M & \underline{51.34} \scriptsize{$\pm$ 0.16} \\
ResNet18~\cite{he2016deep} & 11.26 M & 50.84 \scriptsize{$\pm$ 0.24} \\
ShuffleNetV2x0.5~\cite{ma2018shufflenet} & 0.35 M & 50.08 \scriptsize{$\pm$ 0.37} \\
WiFlexFormer [ours] & 0.05 M & 49.72 \scriptsize{$\pm$ 0.12} \\
\bottomrule
\end{tabularx}
\vspace{-2mm}
\caption{Cross-receiver gesture recognition performance on the \textbf{Widar3.0-G6 dataset} using \textbf{DFS features}. All models are trained on data from receivers 1-3 with an 8:2 training-validation split and tested on receivers 4-6. Amplitude features from all 30 subcarriers are used as input. Results are presented as mean and standard deviation across three independent runs with random initialization.}
\label{tableWIDAR3G6DFS}
\vspace{-5mm}
\end{table}

For amplitude features, the highest accuracy is achieved using all subcarriers (None). Although uniform sampling of every 4th subcarrier (U4) and band-restricted sampling using eight bands with four subcarriers per band (B8-4) yield comparable results, the reduction in preprocessing and inference time for amplitude features is negligible, making sub-sampling unnecessary. For DFS features, the best accuracy is also obtained using all subcarriers, which is to be expected. However, subcarrier sub-sampling strategies, such as uniform sampling of every 2nd subcarrier (U2) or band-restricted sub-sampling with four bands and four subcarriers per band (B4-4), achieve similar accuracy while reducing the number of STFT computations to half and one-fourth, respectively. These strategies are a potential way to further reduce inference time, especially when using DFS features.

\vspace{-1mm}
\section{Limitations and Future Work}
\vspace{-1mm}
The lightweight design of \textit{WiFlexFormer} offers advantages for real-time inference. With its low parameter count and an inference time of approximately 10 ms (Nvidia Jetson Orin Nano), it is well-suited for deployment on embedded devices, particularly in applications requiring high packet rates (e.g., 1-2 kHz for gesture recognition)~\cite{HernandezWiFiOnTheEdgeSurvey}. However, its performance in real-world scenarios and across different edge hardware setups remains to be explored. 

Moreover, due to its small size and fast training capabilities, \textit{WiFlexFormer} is well-suited for techniques like test-time training, allowing rapid adaptation to new WiFi domains~\cite{sun2020test}. This would be especially useful in dynamic environments like smart homes, enabling fine-tuning on-the-fly with minimal computational overhead and system downtime.

While this work focused on comparing relative performance, \textit{WiFlexFormer}'s accuracy could be improved with techniques like ensemble models~\cite{dvornik2019diversity}, which remain computationally feasible given its low parameter count. Pre-training and data augmentation, though not used here to avoid bias and variability, could further be leveraged to enhance absolute performance and generalization, especially in cross-domain scenarios~\cite{Hou2024RFBoostUA}.

A limitation of this work is the evaluation's focus on two datasets: the 3DO dataset with three macroscopic activities and the Widar3.0-G6 dataset with six gestures. Future research should expand evaluation to diverse environments, hardware setups, and multi-person scenarios to better assess \textit{WiFlexFormer}'s generalization and applicability in various WiFi-based sensing domains.

\definecolor{yellow}{HTML}{F7FDC5}
\definecolor{blue}{HTML}{95F0D8}
\definecolor{purple}{HTML}{660099}
\definecolor{cyan}{HTML}{58CFA7}

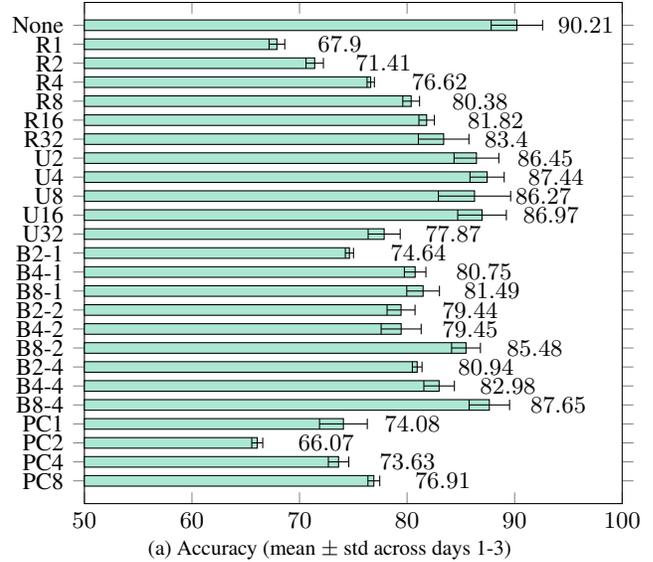
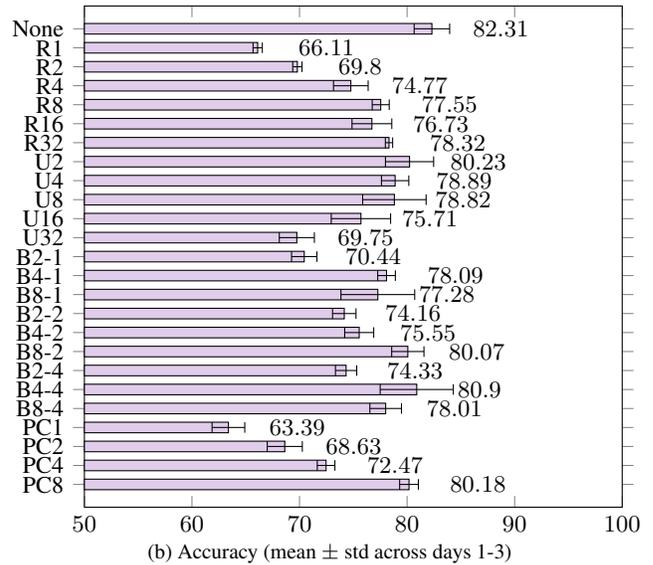
\begin{figure}[t!]
    \centering
    \hspace*{-3mm}
    \begin{subfigure}[t]{0.5\textwidth}
        \centering
        \begin{tikzpicture}
        \begin{axis}[
            xbar,
            symbolic y coords={None,R1,R2,R4,R8,R16,R32,U2,U4,U8,U16,U32,B2-1,B4-1,B8-1,B2-2,B4-2,B8-2,B2-4,B4-4,B8-4,PC1,PC2,PC4,PC8},
            ytick=data,
            xmin=50,
            xmax=100,
            xlabel style={yshift=0.5ex,font=\small},
            x tick label style={/pgf/number format/.cd, fixed, fixed zerofill, precision=0, /tikz/.cd,font=\small},
            y tick label style={/pgf/number format/.cd,fixed,fixed zerofill,precision=0,/tikz/.cd,font=\small},
            bar width=4pt,
            enlarge y limits=0.05,
            nodes near coords,
            every node near coord/.append style={font=\small, xshift=12pt},
            height=8.25cm,
            width=\textwidth,
            y dir=reverse,
            error bars/x dir=both,
            error bars/x explicit,
        ]
        \addplot[xbar, fill=white!50!cyan, error bars/.cd, x explicit] coordinates {
            (90.21,None) +- (2.39,2.39) 
            (67.90,R1) +- (0.73,0.73) 
            (71.41,R2) +- (0.81,0.81) 
            (76.62,R4) +- (0.34,0.34) 
            (80.38,R8) +- (0.78,0.78)
            (81.82,R16) +- (0.72,0.72) 
            (83.40,R32) +- (2.35,2.35) 
            (86.45,U2) +- (2.08,2.08) 
            (87.44,U4) +- (1.59,1.59) 
            (86.27,U8) +- (3.36,3.36) 
            (86.97,U16) +- (2.26,2.26) 
            (77.87,U32) +- (1.49,1.49) 
            (74.64,B2-1) +- (0.38,0.38) 
            (80.75,B4-1) +- (1.00,1.00)
            (81.49,B8-1) +- (1.52,1.52) 
            (79.44,B2-2) +- (1.30,1.30) 
            (79.45,B4-2) +- (1.86,1.86)
            (85.48,B8-2) +- (1.34,1.34)
            (80.94,B2-4) +- (0.45,0.45)
            (82.98,B4-4) +- (1.42,1.42)
            (87.65,B8-4) +- (1.88,1.88)
            (74.08,PC1) +- (2.22,2.22) 
            (66.07,PC2) +- (0.51,0.51) 
            (73.63,PC4) +- (0.95,0.95) 
            (76.91,PC8) +- (0.55,0.55) 
        };
        \end{axis}
        \end{tikzpicture}
        \vspace{-1mm}
        \caption{Accuracy (mean $\pm$ std across days 1-3)}
        \label{figSSAmplitude}
    \end{subfigure}
    \vspace{2mm}
    
    \centering
    \hspace*{-3mm}
    \begin{subfigure}[t]{0.5\textwidth}
        \centering
        \begin{tikzpicture}
        \begin{axis}[
            xbar,
            symbolic y coords={None,R1,R2,R4,R8,R16,R32,U2,U4,U8,U16,U32,B2-1,B4-1,B8-1,B2-2,B4-2,B8-2,B2-4,B4-4,B8-4,PC1,PC2,PC4,PC8},
            ytick=data,
            xmin=50,
            xmax=100,
            xlabel style={yshift=0.5ex,font=\small},
            x tick label style={/pgf/number format/.cd, fixed, fixed zerofill, precision=0, /tikz/.cd,font=\small},
            y tick label style={/pgf/number format/.cd,fixed,fixed zerofill,precision=0,/tikz/.cd,font=\small},
            bar width=4pt,
            enlarge y limits=0.05,
            nodes near coords,
            every node near coord/.append style={font=\small, xshift=12pt},
            height=8.25cm,
            width=\textwidth,
            y dir=reverse,
            error bars/x dir=both,
            error bars/x explicit,
        ]
        \addplot[xbar, fill=white!80!purple, error bars/.cd, x explicit] coordinates {
            (82.31,None) +- (1.65,1.65) 
            (66.11,R1) +- (0.43,0.43) 
            (69.80,R2) +- (0.43,0.43) 
            (74.77,R4) +- (1.61,1.61)  
            (77.55,R8) +- (0.79,0.79) 
            (76.73,R16) +- (1.85,1.85) 
            (78.32,R32) +- (0.34,0.34) 
            (80.23,U2) +- (2.24,2.24) 
            (78.89,U4) +- (1.27,1.27) 
            (78.82,U8) +- (2.95,2.95) 
            (75.71,U16) +- (2.76,2.76) 
            (69.75,U32) +- (1.63,1.63) 
            (70.44,B2-1) +- (1.18,1.18) 
            (78.09,B4-1) +- (0.82,0.82) 
            (77.28,B8-1) +- (3.43,3.43) 
            (74.16,B2-2) +- (1.09,1.09) 
            (75.55,B4-2) +- (1.35,1.35) 
            (80.07,B8-2) +- (1.50,1.50) 
            (74.33,B2-4) +- (0.99,0.99) 
            (80.90,B4-4) +- (3.40,3.40) 
            (78.01,B8-4) +- (1.47,1.47) 
            (63.39,PC1) +- (1.52,1.52) 
            (68.63,PC2) +- (1.63,1.63) 
            (72.47,PC4) +- (0.82,0.82) 
            (80.18,PC8) +- (0.87,0.87)  
        };
        \end{axis}
        \end{tikzpicture}
        \vspace{-1mm}
        \caption{Accuracy (mean $\pm$ std across days 1-3)}
        \label{figSSDFS}
    \end{subfigure}
    \vspace{-3mm}
    \caption{Comparative analysis of subcarrier selection strategies for (a) \textbf{amplitude features}, and (b) \textbf{DFS features} using the \textbf{3DO dataset}. Strategies include: (1) None: use of all subcarriers; (2) R$n$: random selection of $n$ subcarriers; (3) U$n$: uniform selection of every $n$th subcarrier; (4) B$n$-$m$: division into $n$ subcarrier bands with random selection of $m$ subcarriers from each band; and (5) PC$n$: selection of the first $n$ principal components.}
    \label{figSS}
    \vspace{-4mm}
\end{figure}

\section{Conclusion}
\vspace{-1mm}
In this work, we proposed \textit{WiFlexFormer}, a highly efficient Transformer-based architecture designed for processing WiFi CSI features. Our comprehensive evaluation on publicly available WiFi datasets, assessing HAR performance using amplitude and DFS features, as well as inference time, demonstrates that \textit{WiFlexFormer} compares favorably against state-of-the-art architectures. With only approximately 50k parameters and an inference time of around 10 ms (Nvidia Jetson Orin Nano), \textit{WiFlexFormer} achieves a significant three orders of magnitude reduction in parameter count, making it particularly well-suited for on-device inference in real-time WiFi-based sensing applications at the edge. Additionally, we investigated the effectiveness of subcarrier sub-sampling strategies and identified uniform and band-restricted random sub-sampling as potential ways to further reduce computational complexity. The 3DO dataset used in our evaluation, along with a PyTorch implementation of \textit{WiFlexFormer}, is made publicly available for further research and development.

{\small
\bibliographystyle{ieeenat_fullname}
\bibliography{main}
}

\end{document}